\documentclass{article}

\usepackage[main, final]{template}

\usepackage[utf8]{inputenc} 
\usepackage[T1]{fontenc}    
\usepackage{hyperref}       
\hypersetup{hidelinks}
\usepackage{url}            
\usepackage{booktabs}       
\usepackage{amsfonts}       
\usepackage{nicefrac}       
\usepackage{microtype}      
\usepackage{xcolor}         
\usepackage{amsmath}
\usepackage{xspace}
\usepackage{epsfig}
\usepackage{graphicx}
\usepackage{amssymb}
\usepackage{bm}
\usepackage{mathrsfs}
\usepackage{pifont}
\usepackage{float}
\usepackage{multicol}
\usepackage{multirow}
\usepackage{adjustbox}
\usepackage{wrapfig}
\usepackage{array}
\usepackage[table]{xcolor}
\usepackage{algorithm}
\usepackage{algpseudocode}
\newcommand{\eg}{\emph{e.g.}}

\usepackage[capitalize]{cleveref}
\crefname{section}{Sec.}{Secs.}
\Crefname{section}{Section}{Sections}
\Crefname{table}{Table}{Tables}
\crefname{table}{Tab.}{Tabs.}
\crefname{algorithm}{Alg.}{Algs.}
\Crefname{algorithm}{Algorithm}{Algorithms}

\usepackage{titlesec}
\titlespacing*{\section}{0pt}{6pt plus 1pt minus 1pt}{3pt plus 1pt minus 1pt}
\titlespacing*{\subsection}{0pt}{5pt plus 1pt minus 1pt}{2pt plus 1pt minus 1pt}
\titlespacing*{\subsubsection}{0pt}{4pt plus 1pt minus 1pt}{2pt plus 1pt minus 1pt}
\titlespacing*{\paragraph}{0pt}{3pt plus 1pt minus 1pt}{0.5em}

\newcommand{\method}{{HorizonDrive}\xspace}
\newcommand{\equalcontrib}{\textsuperscript{*}}
\newcommand{\corrauthor}{\textsuperscript{†}}
\newcommand{\projectlead}{\textsuperscript{‡}}
\title{HorizonDrive: Self-Corrective Autoregressive World Model for Long-horizon Driving Simulation}

\author{%
\textbf{Conglang Zhang}$^1\equalcontrib$ \quad \textbf{Yifan Zhan}$^2\equalcontrib$ \quad \textbf{Qingjie Wang}$^3$ \quad \textbf{Zhanpeng Ouyang}$^3$ \quad \textbf{Yu Li}$^4$ 
\\ 
\quad \textbf{Zihao Yang}$^5$ \quad \textbf{Xiaoyang Guo}$^6$  \quad \textbf{Weiqiang Ren}$^3$ \quad \textbf{Qian Zhang}$^3$ \quad \textbf{Zhen Dong}$^1$
\\
 \textbf{Yinqiang Zheng}$^2$ \quad \textbf{Wei Yin}$^3\projectlead$\quad \textbf{Zhengqing Chen}$^3\corrauthor$
\\\\
$^1$Wuhan University \quad $^2$The University of Tokyo \quad $^3$ Horizon Robotics \quad $^4$ Tsinghua University  \\ \quad $^5$ University of Science and Technology of China \quad $^6$ The Chinese University of Hong Kong
\\
\textsuperscript{*}Equal contribution \quad
\textsuperscript{‡}Project lead \quad
\textsuperscript{†}Corresponding author
}

\begin{document}

\maketitle

\begin{abstract}
Closed-loop driving simulation requires real-time interaction beyond short offline clips, pushing current driving world models toward autoregressive (AR) rollout. Existing AR distillation approaches typically rely on frame sinks or student-side degradation training.
The former transfers poorly to driving due to fast ego-motion and rapid scene changes, while the latter remains bounded by the teacher's single-pass output length and thus provides only a limited supervision horizon.
A natural question is: \textit{can the teacher itself be extended via AR rollout to provide unbounded-horizon supervision at bounded memory cost?}
The key difficulty is that a standard teacher drifts under its own predictions, contaminating the supervision it provides.
Our key insight is to make the teacher rollout-capable, ensuring reliable supervision from its own AR rollouts.
This is instantiated as \method, an anti-drifting training-and-distillation framework for AR driving simulation.
First, scheduled rollout recovery (SRR) trains the base model to reconstruct ground-truth future clips from prediction-corrupted histories, yielding a teacher that remains stable across long AR rollouts.
Second, the rollout-capable teacher is extended via AR rollout, providing long-horizon distribution-matching supervision under bounded memory, while a short-window student aligns to it with teacher rollout DMD (TRD) for efficient real-time deployment.
\method natively supports minute-scale AR rollout under bounded memory; on nuScenes, \method reduces FID by 52\% and FVD by 37\%, and lowers ARE and DTW by 21\% and 9\% relative to the strongest long-horizon streaming baselines, while remaining competitive with single-pass driving video generators.
See project page at \href{https://zcliangyue.github.io/HorizonDrive}{\textcolor{blue}{https://zcliangyue.github.io/HorizonDrive}}.
\end{abstract}

\section{Introduction}

\begin{figure}[t]
\centering
\includegraphics[width=1\linewidth]{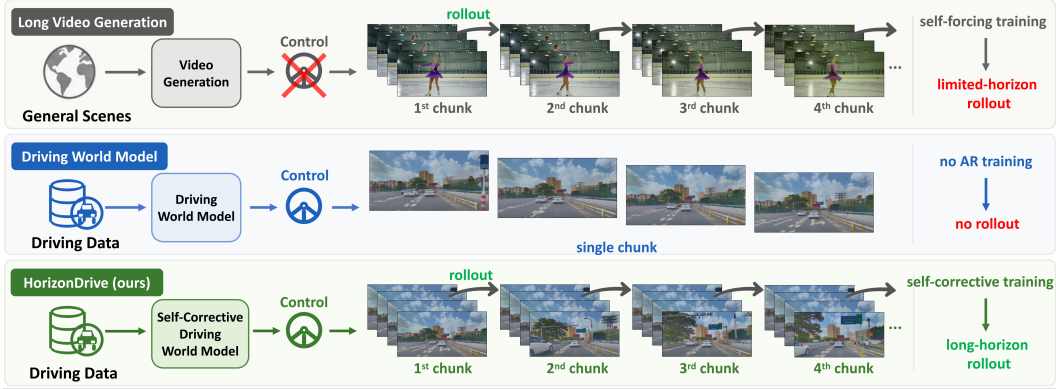}
\vspace{-3ex}
\caption{
\textbf{Comparison with general long-video generators and driving world models.}
General long-video generators can roll out but lack driving-specific control and suffer from drift, while existing driving world models cannot roll out autoregressively.
HorizonDrive enables both action-controllable generation and stable long-horizon AR rollout, supporting real-time interactive driving simulation.
}
\label{fig:teaser}
\end{figure}

Driving world models aim to serve as closed-loop simulation testbeds, where generated environments respond to an agent's actions in real time.
This requires long-horizon autoregressive (AR) rollout, in which the model repeatedly generates short future chunks, receives updated controls, and feeds its own predictions back as context.
Such deployment induces exposure bias: each chunk is conditioned on imperfect generated frames rather than ground-truth observations, causing small errors to recursively compound into visual artifacts, geometric inconsistency, and semantic drift.
As a result, even state-of-the-art driving world models (\eg,~\citet{nvidia2025cosmosdrivedreams, gao2025magicdrive}) degrade quickly, far short of the minute-scale horizons needed for meaningful closed-loop evaluation.


Recent work has begun to address this gap.
Self-Forcing~\citep{huang2025self} trains an autoregressive student from a standard teacher diffusion model by conditioning each chunk on the student's previous outputs rather than ground truth, exposing it to realistic rollout errors.
While this improves rollout stability, the corrective signal is still limited to the teacher's fixed single-pass generation window.
Extending this window is impractical, as the attention cost of DiT grows quadratically with sequence length and quickly exceeds memory limits.
Consequently, the supervision horizon remains capped by the teacher's single-pass capacity, making long-horizon AR distillation hard to scale.


A natural question then arises: \textit{can the teacher itself be extended to arbitrary horizons through AR rollout, providing unbounded supervision at constant memory cost?}
We show that this is possible once the teacher is first made rollout-capable: by remaining stable under its own predicted context, the teacher breaks the single-pass horizon barrier and extends supervision through AR rollout.
Based on this insight, we propose \textbf{\method}, a two-stage anti-drifting distillation framework for real-time autoregressive driving simulation.


In the first stage, \textit{Scheduled Rollout Recovery (SRR)}, we convert a standard driving world model into a rollout-capable teacher.
This departs from conventional anti-drifting paradigms, which mainly regularize the deployed generator; instead, SRR stabilizes the supervisor so that long-horizon distillation is no longer confined to a clean, single-pass teacher window.
SRR trains the teacher to recover clean future dynamics from prediction-corrupted histories, with a joint local-and-global schedule that smooths pred-to-GT transitions and shifts training from severe late-rollout drift to finer early-rollout correction.
This enables stable repeated fixed-window rollout without relying on frame sinks or increasing single-pass memory cost.

In the second stage, we introduce \textit{Teacher Rollout DMD (TRD)} to transfer long-horizon rollout behavior into a real-time student.
The rollout-capable teacher provides stable long-horizon supervision through repeated fixed-window rollout, and TRD aligns the student to this teacher trajectory in a chunk-wise manner.
This asymmetric design allows the teacher to generate longer chunks with more denoising steps, while the student generates shorter chunks with fewer steps for faster and finer-grained interaction.
By rolling out multiple short student chunks to match the same teacher horizon, TRD enables an efficient student to inherit the teacher's long-horizon stability.



Overall, \method yields a short-chunk student that natively supports long-horizon AR rollout for interactive driving simulation.
As summarized in Fig.~\ref{fig:teaser}, it bridges general long-video generation and driving world modeling by combining action-controllable generation with stable AR rollout.
The student rolls out indefinitely with 10-frame chunks under bounded memory and can be extended to minute-scale horizons, while supporting multi-condition control (text, HD map, bounding boxes, and ego action).
On nuScenes (where the per-clip length is bounded to $\sim$20\,s by the dataset), \method reduces FID by 52\% and FVD by 37\%, and lowers ARE and DTW by 21\% and 9\% over the strongest long-horizon streaming baselines, while remaining competitive with single-pass driving video generators on short-clip evaluation; minute-scale rollout is further demonstrated qualitatively in the appendix.

Our contributions are threefold:
(1) We identify rollout-capable teaching as the missing prerequisite for scalable long-horizon distillation and propose SRR to turn a standard driving world model into a stable AR teacher.
(2) We introduce TRD, which distills teacher rollout trajectories into a short-chunk, few-step student under bounded memory.
(3) We provide a systematic evaluation on nuScenes, demonstrating that teacher-side rollout stabilization and chunk-wise rollout distillation substantially improve long-horizon visual quality and spatio-temporal consistency over strong streaming baselines.

\section{Related Works}

\paragraph{Conditional Video Generation for Autonomous Driving.}
Recent progress in video synthesis has been largely enabled by diffusion-based generative models~\citep{peebles2023scalable, brooks2024video, gao2025seedance, runwayml2024introducing, yang2024cogvideox, kong2024hunyuanvideo, ma2025step, wan2025wan, team2025longcat}.
Autonomous driving requires generation models to be both visually realistic and precisely controllable by structured scene conditions and driving behaviors.
Existing methods often introduce geometric priors, including HD maps, 3D boxes, occupancy, and object-level spatial structures, to guide scene composition and agent placement~\citep{wen2024panacea, wang2024driving, gao2024magicdrive3d, zhao2025drivedreamer, gao2025magicdrive}.
Other methods use ego-motion or trajectory cues to control ego-vehicle behavior~\citep{hu2023gaia, lu2024wovogen, gao2024vista, zhang2025epona, russell2025gaia, zhan2026composing}.
Yet, most existing methods are limited by short generation horizons or high inference cost, restricting their use in closed-loop simulation.

\paragraph{Long Video Generation.}
Long-horizon video generation remains challenging, as models trained on short clips often accumulate temporal drift when extended to longer durations.
Existing methods address this issue through training-free temporal extension, rollout-aware training, or inference-time stabilization.
\citet{qiu2023freenoise} and~\citet{kim2024fifo} modify noise scheduling or reuse noisy contexts to extend short-video models.
\citet{chen2024diffusion} and~\citet{liu2025rolling} train with frame-wise independent noise to simulate corrupted histories, enabling autoregressive generation at inference time.
Other efforts explore next-frame prediction with inverted sampling~\citep{zhang2025packing}, causal rollout training~\citep{huang2025self}, error-bank mechanisms~\citep{guo2025end, li2025stable}, keyframe sinking~\citep{huang2024context, xiao2025captain, zhou2024storydiffusion}, test-time training~\citep{dalal2025one}, and multi-shot generation~\citep{cai2025mixture}.
Despite these advances, existing methods still suffer from drift beyond the training horizon.

\paragraph{Efficient Video Generation via Diffusion Distillation.}

Diffusion-based video models rely on iterative denoising, making inference cost a key obstacle to real-time generation.
Recent methods accelerate sampling by distilling a multi-step teacher into a few-step student, using distribution matching~\citep{yin2024one, yin2024improved, liu2025decoupled, wu2026diversity} or consistency-style objectives~\citep{song2023consistency, geng2024consistency, geng2025mean, zheng2025large}.
\citet{huang2025self} further introduces rollout training to reduce the training-inference gap.
However, distilled models can still accumulate errors during long-horizon autoregressive rollouts, limiting their robustness in closed-loop simulation.
Our method performs teacher-rollout DMD, transferring long-horizon behavior from a stronger teacher to a short-chunk student while preserving efficient inference.

\section{Preliminary}

\paragraph{Diffusion models.} In generative modeling, a diffusion process~\citep{rombach2022high, esser2024scaling, blattmann2023stable, brooks2024video, zheng2024open} provides a continuous transformation of noise into structured data through a differential equation.
This process models the evolution of latent variables, where each time step is controlled by a vector field \( v_\Theta \), defined as
\begin{equation}
\label{equ:flow_matching}
    \mathrm{d}z_{(t)} = v_\Theta(z_{(t)}, t)\,\mathrm{d}t, \quad t \in [0, 1].
\end{equation}

The latent variables \( z_{(t)} \) evolve from random Gaussian noise \( \boldsymbol{\epsilon} \) at \( t = 1 \) to the data distribution at \( t = 0 \).
The intermediate latent states are constructed as
\begin{equation}
\label{equ:states}
    z_{(t)} = \sigma_t\,z_{(0)} + (1 - \sigma_t)\,\boldsymbol{\epsilon},
\end{equation}

where \( \sigma_t \) is a predefined noise schedule. The objective for recent flow matching formulation~\citep{albergo2022building,
lipman2022flow, liu2022flow} is defined by the \( v \)-prediction loss as
\begin{equation}
\label{l_cfm}
    \mathcal{L}_{CFM} =
    \mathbb{E}_{\boldsymbol{\epsilon} \sim \mathcal{N}(\mathbf{0}, \mathbf{I})}
    \left\|
        v_\Theta(z_{(t)}, t) - (z_{(0)} - \boldsymbol{\epsilon})
    \right\|_2^2.
\end{equation}

\paragraph{Distribution matching distillation.} Due to the slow inference speed of diffusion models, Distribution Matching Distillation (DMD)~\citep{yin2024one} is applied to enhance real-time performance.
DMD trains a few-step student model to match the teacher's output distribution, thus reducing the inference overhead while maintaining high-quality generation.
The loss gradient is defined as
\begin{equation}
\label{equ:dmd}
    \nabla_\theta \mathcal{L}_{DMD} = \mathbb{E}_\tau \left[ - \left( s_{\text{real}}(z_{(\tau)}) - s_{\text{fake}}(z_{(\tau)}) \right) \right] \frac{\partial G_\theta}{\partial \theta},
\end{equation}
where $\tau$ is renoised level.
This ensures similar generative quality between student and teacher models.

\section{Methods}
\label{sec: methods}

\begin{figure}[t]
\centering
\includegraphics[width=1\linewidth]{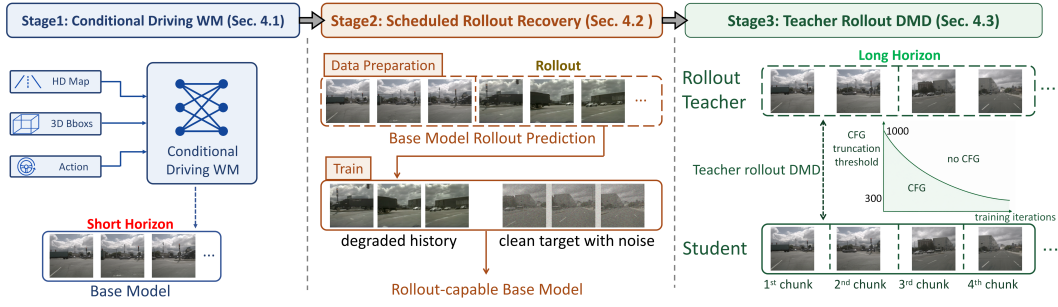}
\vspace{-3ex}
\caption{\textbf{Overview of \method framework.}
We first train a conditional driving world model, then improve its autoregressive stability through scheduled rollout recovery, and finally distill long-horizon teacher rollouts into a few-step, short-chunk student via teacher-rollout DMD.}
\label{fig:pipeline}
\end{figure}

\paragraph{Problem formulation.} 

We aim to build a real-time interactive autoregressive (AR) driving world model in a video-VAE latent space, where \(\mathbf{z}_{1:T}\) denotes the AR context window and \(\mathbf{c}_{T+1:T+K}\) denotes future driving controls over the next $K$ frames.
At each step, the model predicts the next $K$-frame chunk as
\begin{equation}
\label{eq: ar_generation}
\hat{\mathbf{z}}_{T+1:T+K}
\sim
p_\theta
\left(
\mathbf{z}_{T+1:T+K}
\mid
\mathbf{z}_{1:T}, \mathbf{c}_{T+1:T+K}
\right),
\end{equation}
and appends it to the history buffer for the next step.

To realize this AR process, we instantiate \method through three stages, as illustrated in~\cref{fig:pipeline}.
First,~\cref{subsec: Base model training} trains a standard driving world model with robust multi-condition control, which serves as the foundation for subsequent autoregressive rollout.
Second,~\cref{subsec: Scheduled rollout recovery} applies scheduled rollout recovery to convert this standard model into a rollout-capable teacher, enabling it to maintain stable generation under its own predicted histories.
Finally,~\cref{subsec: teacher rollout DMD training} performs teacher-rollout DMD, where the long-chunk, multi-step teacher provides rollout-consistent supervision for a short-chunk, few-step student.

\subsection{Conditional driving world model training}
\label{subsec: Base model training}

The backbone of our conditional driving world model is a pretrained text-to-video (T2V) diffusion transformer with full bidirectional attention.
To turn it into a video continuation model without modifying its architecture, we partition each training clip into a $T$-frame condition window and a $K$-frame generation chunk, and assign different noise levels to the two parts: condition latents \(\mathbf{z}_{1:T}\) are kept clean with noise level $t{=}0$, while chunk latents \(\mathbf{z}_{T+1:T+K}\) are noised and supervised with the flow-matching loss in~\cref{l_cfm}.

To enable conditional control, we follow disentangled control~\citep{nvidia2025cosmosdrivedreams, russell2025gaia, zhan2026composing}: spatial scene structure (HD map, bounding boxes) is injected as additive layout tokens via a zero-initialized projector, while the residual ego-action signal $\bm{a}{=}(\Delta x, \Delta y, \Delta\text{yaw})$ is injected through AdaLN-style gating~\citep{peebles2023scalable}. Architectural details and the exact injection forms are deferred to~\cref{appx:control_modules}.

After training, we obtain a standard controllable driving world model, denoted as $\mathcal{G}_0$.
At inference, $\mathcal{G}_0$ rolls out long sequences via a sliding window: the most recent $T$ latents form the next condition, a new $K$-frame chunk is generated via~\cref{eq: ar_generation}, and the context window is then shifted forward by $K$ frames for the next AR step.
However, since $\mathcal{G}_0$ is trained only with clean ground-truth histories, direct autoregressive rollout suffers from exposure bias.

\subsection{SRR: scheduled rollout recovery training}
\label{subsec: Scheduled rollout recovery}

To obtain a rollout-capable base model, instead of training on clean ground-truth clips, we sample segments from the rollout trajectory of $\mathcal{G}_0$ as degraded training conditions.
Starting from a ground-truth latent history with length $T$, we roll out $\mathcal{G}_0$ for $N$ autoregressive steps. 
At each step, the model predicts a chunk of $K$ future latents and appends them to a fixed-length history buffer of size $T$:
\begin{equation}
    \hat{\mathbf{z}}_{s_n+1:s_n+K}
    =
    \mathcal{G}_0
    \left(
    \hat{\mathbf{z}}_{s_n-T+1:s_n},
    \mathbf{c}_{s_n+1:s_n+K}
    \right),
    \quad n=1,\ldots,N ,
\end{equation}
where $s_n = T + (n-1)K$. This produces an $N$-step rollout trajectory
\(\hat{\mathbf{z}}_{T+1:T+NK}\), in which prediction errors accumulate across AR steps.

For a sampled starting index $s$ for generation, we replace the conditioning history with rollout predictions while keeping the supervision target unchanged:
\begin{equation}
    \tilde{\mathbf{z}}_{s-T+1:s}
    =
    \hat{\mathbf{z}}_{s-T+1:s},
    \qquad
    \mathbf{z}^{\star}_{s+1:s+K}
    =
    \mathbf{z}_{s+1:s+K}.
\end{equation}

\begin{figure}[t]
\centering
\includegraphics[width=1\linewidth]{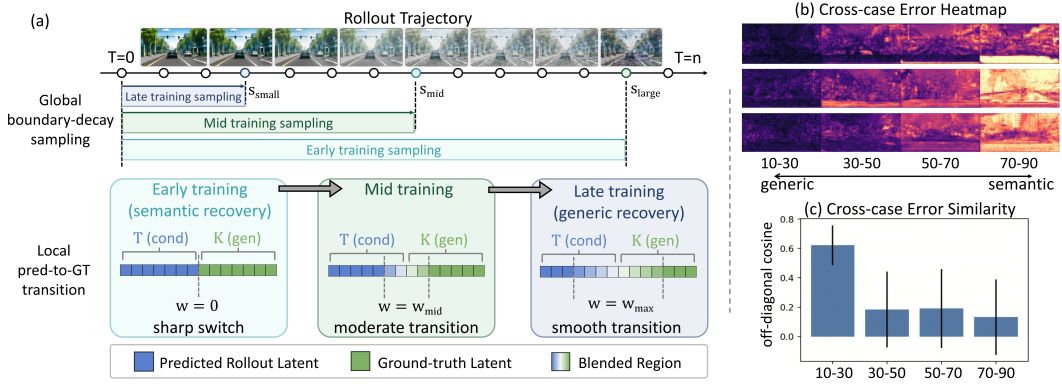}
\vspace{-3ex}
\caption{\textbf{Details of scheduled rollout recovery.}
(a) Boundary-decay sampling gradually shifts training from late, semantically drifted rollout regions to earlier, more generic degradation, while pred-to-GT transition smooths the recovery target.
(b) Error heatmaps reveal stronger semantic corruption at later rollout intervals.
(c) Cross-case similarity shows that earlier errors are more consistent, supporting the proposed joint local-and-global schedule.}
\label{fig:stage1}
\end{figure}

As shown in~\cref{fig:stage1} (a), we use two complementary schedules to make recovery training both smooth and progressive: 
a local pred-to-GT transition schedule within each sampled segment, and a global boundary-decay sampling schedule along the rollout trajectory.

\paragraph{Local pred-to-GT transition.}
Directly switching from the predicted history to the ground-truth future creates an abrupt temporal boundary.
To smooth this transition, we introduce a pred-to-GT blending window of radius $w$ around the generation boundary $s$.
The final mixed training sequence $\bar{\mathbf{z}}_{i}$ is constructed as
\begin{equation}
\bar{\mathbf{z}}_{i}
=
\begin{cases}
\tilde{\mathbf{z}}_{i}, 
& s-T+1 \le i \le s-w, \\[2pt]
\alpha_i \tilde{\mathbf{z}}_{i}
+
(1-\alpha_i)\mathbf{z}^{\star}_{i},
& s-w+1 \le i \le s+w, \\[2pt]
\mathbf{z}^{\star}_{i},
& s+w+1 \le i \le s+K,
\end{cases}
\end{equation}
where $\alpha_i$ linearly decreases from $1$ to $0$ within the transition window.
This constructs a continuous temporal bridge from rollout-degraded latents to ground-truth latents around the boundary.

We further schedule the blending radius during training.
At the beginning, we set $w=0$, which exposes the model to a sharp boundary and encourages direct recovery from large rollout-induced deviations.
As training progresses, $w$ is gradually increased, expanding the transition region and shifting the task toward smoother, fine-grained correction.

\paragraph{Global boundary-decay sampling.}
We further schedule the sampled generation boundary $s$ along the rollout trajectory.
Here, the ranges $10$--$30$, $30$--$50$, $50$--$70$, and $70$--$90$ in~\cref{fig:stage1} (b)(c) denote different boundary intervals of $s$ from early to late rollout positions.
The visualization reveals two complementary properties of rollout errors.
First, the cross-case error heatmaps become progressively stronger as $s$ moves toward later boundaries, indicating that longer autoregressive rollout leads to more severe accumulated degradation and semantic drift.
Second, the off-diagonal cosine similarity is highest in the early boundary interval, showing that early-stage errors are more shared across cases, while later-stage semantic errors are more case-specific.
Motivated by these observations, we adopt a boundary-decay curriculum: training starts from larger $s$, where the model is forced to recover from severe long-horizon semantic drift, and $s$ is gradually decayed toward earlier rollout positions.
This schedule first builds robustness to accumulated semantic failures and then refines the model on more generic, cross-case consistent rollout degradation.

Through scheduled rollout recovery (SRR), the initial conditional model 
$\mathcal{G}_0$ is transformed into a rollout-capable base model 
$\mathcal{G}_{\mathrm{roll}}$, which learns to correct prediction-induced errors under realistic AR conditions and provides reliable long-horizon supervision for subsequent distillation.

\subsection{TRD: teacher rollout DMD training}
\label{subsec: teacher rollout DMD training}

Despite its improved rollout stability, $\mathcal{G}_{\mathrm{roll}}$ remains constrained by the high inference cost of multi-step diffusion sampling, making it impractical for real-time interactive simulation.
We therefore distill it into a faster student through teacher rollout DMD training (TRD).
Both the teacher $\mathcal{G}^{\mathcal{T}}_{\mathrm{roll}}$ and the student $\mathcal{G}^{\mathcal{S}}_{\mathrm{roll}}$ are initialized from $\mathcal{G}_{\mathrm{roll}}$.
The teacher is kept frozen, while only the student parameters $\phi$ are updated.
They share the same conditioning window length $T$, but use different chunk sizes during generation.
The teacher uses a longer chunk size $K^{\mathcal{T}}$, while the student uses a shorter chunk size $K^{\mathcal{S}}$, with $K^{\mathcal{T}}>K^{\mathcal{S}}$.

For long-horizon supervision beyond a single teacher chunk, we do not increase the teacher generation length.
Instead, the teacher itself performs AR rollout with the fixed $(T,K^{\mathcal{T}})$ window.
This design keeps the teacher memory footprint fixed.
The student also rolls out autoregressively with the fixed $(T,K^{\mathcal{S}})$ window:
\begin{equation}
\hat{\mathbf{z}}^{\mathcal{S}}_{s_n+1:s_n+K^{\mathcal{S}}}
=
\mathcal{G}^{\mathcal{S}}_{\mathrm{roll},\phi}
\left(
\hat{\mathbf{z}}^{\mathcal{S}}_{s_n-T+1:s_n},
\mathbf{c}_{s_n+1:s_n+K^{\mathcal{S}}}
\right),
\quad n=1,\ldots,N ,
\end{equation}
where $\phi$ denotes the student parameters.
Whenever the accumulated student rollout covers a teacher-length interval $K^{\mathcal{T}}$, we apply the DMD gradient to the latest $K^{\mathcal{T}}$ frames and immediately backpropagate.
The frozen teacher rolls out over the same interval with its fixed $(T,K^{\mathcal{T}})$ window to provide the distribution-matching direction.
This step is repeated along the trajectory, transferring the teacher's corrective ability to the student without accumulating memory across rollout steps.

Moreover, we adopt a noise-truncated CFG strategy. Standard DMD typically uses CFG to strengthen the teacher's gradient guidance, but this tends to cause oversaturation during video rollout. Following Decoupled DMD~\citep{liu2025decoupled}, we restrict CFG to noise levels below a threshold $\tau_{\mathrm{th}}$. We further schedule $\tau_{\mathrm{th}}$ to decay during training, shifting the optimization focus from early conditional controllability to late-stage visual refinement.

Extending the standard DMD loss in~\eqref{equ:dmd}, the TRD gradient is written as
\begin{equation}
\begin{split}
    \nabla_{\phi}\mathcal{L}_{\mathrm{TRD}}
    =
    \mathbb{E}_{\tau}
    \Bigg[
    -\Bigg(
    &\underbrace{
    s_{\mathrm{cond}}^{\mathrm{real}}
    \left(z_{(\tau)}\right)
    -
    s_{\mathrm{cond}}^{\mathrm{fake}}
    \left(z_{(\tau)}\right)
    }_{\mathrm{Distribution\, Matching}}
    \\
    &+
    \underbrace{\mathbf{1}_{\{\tau \leq \tau_{\mathrm{th}}\}}
    (\alpha-1)
    \left(
    s_{\mathrm{cond}}^{\mathrm{real}}
    \left(z_{(\tau)}\right)
    -
    s_{\mathrm{uncond}}^{\mathrm{real}}
    \left(z_{(\tau)}\right)
    \right)
    }_{\mathrm{Noise\text{-}truncated\, CFG}}
    \Bigg)
    \frac{\partial \mathcal{G}^{\mathcal{S}}_{\mathrm{roll},\phi}}{\partial \phi}
    \Bigg],
\end{split}
\end{equation}

Overall, TRD uses a fixed-memory, long-chunk, multi-step teacher to supervise a short-chunk, few-step student along AR rollout trajectories.
By aligning their chunk-wise distributions, the student inherits long-horizon generation behavior while substantially reducing per-chunk denoising latency and inference cost.

\section{Experiments}
\label{sec: experiments}

\subsection{Implementation}
\label{subsec: implementation}

\paragraph{Implementation details.}
\method\ is built on Wan~2.1~1.3B T2V~\citep{wan2025wan} with the disentangled control modules of~\cref{subsec: Base model training}. For TRD, the teacher and student share a context window $T{=}11$, with chunk sizes $K^{\mathcal{T}}{=}40$ (multi-step teacher) and $K^{\mathcal{S}}{=}10$ (4-step student); the student rolls out $N{=}20$ AR steps during training. We use 700 nuScenes~\citep{caesar2020nuscenes} multi-view videos for training and 150 for validation; the per-clip length of $\sim$20\,s is the dataset upper bound and determines our quantitative horizon, while longer minute-scale rollouts are shown qualitatively on a self-collected dataset (\cref{appx:minute}). Full backbone, VAE, dataset, and optimization details are in~\cref{appx:implementation}.

\paragraph{Evaluation metrics.}
We report \textbf{visual quality} via FID and FVD over the full rollout, as well as VBench~\citep{huang2024vbench}. For \textbf{spatio-temporal consistency} (only applicable to methods with driving control), we recover per-frame poses with VGGT~\citep{wang2025vggt} and report \textbf{ARE} (mean geodesic rotation error vs.\ GT) and \textbf{DTW} (dynamic time warping distance between predicted and GT ego-motion trajectories~\citep{keogh2000scaling}); full definitions are in~\cref{appx:metrics}.

\subsection{Comparisons}

\paragraph{Comparison methods.}
To ensure fair comparison across heterogeneous evaluation settings, we compare \method against three groups of baselines on nuScenes; per-group evaluation protocols are detailed in~\cref{appx:baselines}.
\emph{(i) Long-horizon interactive world model frameworks} (top group of~\cref{tab:main_comparison}) are end-to-end frameworks for long-horizon interactive world simulation, including Matrix-Game3~\citep{wang2026matrix}, Helios~\citep{yuan2026helios}, Causal-Forcing~\citep{zhu2026causal}, HY-WorldPlay~\citep{sun2025worldplay}, and LingBot-World~\citep{team2026advancing}; they do not natively accept our driving control signals, so we condition them only on the initial frames and report visual-quality metrics only.
\emph{(ii) Long-horizon streaming video generation methods} (bottom group of~\cref{tab:main_comparison}) are general anti-drifting recipes for streaming video generation, including Self-Forcing~\citep{huang2025self}, Self-Forcing++~\citep{cui2025self}, and LongLive~\citep{yang2025longlive}; we re-train each under our base model and driving control modules so that architecture, base model, data, and conditioning are shared with \method, isolating the long-horizon training framework as the only remaining variable.
\emph{(iii) Driving video generation models} (\cref{tab:driving_specific}) are domain-specific generators (DriveDreamer~\citep{wang2024drivedreamer}, Panacea~\citep{wen2024panacea}, DreamForge~\citep{mei2024dreamforge}, Vista~\citep{gao2024vista}, MagicDrive-V2~\citep{gao2025magicdrive}) that accept driving control signals (actions, HD maps, bounding boxes) but generate only a fixed-length single-pass clip; most report metrics on short clips (8--25 frames), leaving long-video quality unexamined, so we evaluate in two groups---\emph{short-video} and \emph{long-video}---under each method's original generation setting.

\paragraph{World model frameworks produce degraded visuals without driving control.}
The top group of~\cref{tab:main_comparison} reports interactive world model frameworks that were not natively designed for driving and do not accept our control signals. Under our nuScenes long-rollout protocol, their visual fidelity is noticeably limited: FID ranges from 30.53 to 49.07 and FVD from 218.23 to 580.72, and their Image Quality scores (55.55--60.44) remain consistently below \method's 62.50.
This highlights the importance of scene-specific driving control for faithful long-horizon driving simulation.

\paragraph{Self-correction outperforms na\"ive long-horizon streaming methods.}
The bottom group of~\cref{tab:main_comparison} re-trains streaming anti-drifting methods on our base model and driving modules, ensuring identical architecture, data, and control signals---the only remaining variable is the long-horizon training framework itself.
Even so, their visual quality remains consistently worse than \method (FID 2--3$\times$ higher and FVD 1.6--1.7$\times$ higher), and their ego-motion consistency is also weaker on both ARE (3.28--3.78 vs.\ 2.60) and DTW (3.61--6.22 vs.\ 3.27), validating the effectiveness of our SRR--TRD framework.
We attribute the gap to the fact that these baselines rely solely on long-horizon fine-tuning to mitigate drift, yet long tuning alone does not guarantee a rollout-capable teacher: a teacher that cannot reliably perform autoregressive rollout produces degraded trajectories, and distilling from poor supervision yields poor students.
In contrast, SRR strengthens the teacher before distillation begins, raising the ceiling of what the student can learn. Error accumulation is more evident in~\cref{fig:fid_per_chunk}, where the baseline's FID degrades steadily with each autoregressive step while ours remains stable.

\paragraph{Driving-specific methods.}
\Cref{tab:driving_specific} compares \method\ with domain-specific driving generators.
Since evaluation settings differ across methods, we report results from each method's original paper for reference rather than as a strict ranking.
In the short-video group, \method\ with $N{=}1$ generates 21-frame clips that attain the best FVD (84.53) in the group, and its FID (12.54) is competitive with prior short-clip generators while uniquely supporting the full driving control suite (T+M+B+A) at the same chunk size used for long-video evaluation.
In the long-video group, \method\ rolls out $N{=}20$ steps autoregressively and matches MagicDrive-V2's single-pass 241-frame quality (FVD 92.99 vs.\ 94.84, FID 13.82 vs.\ 20.91), suggesting that sequential rollout with few-step denoising can be competitive with many-step, single-pass generation.
Together, these comparisons indicate that \method\ delivers strong long-video quality for autonomous driving video generation.

\begin{figure}[t]
    \centering
    \makebox[\linewidth][c]{
        \includegraphics[width=1.\linewidth]{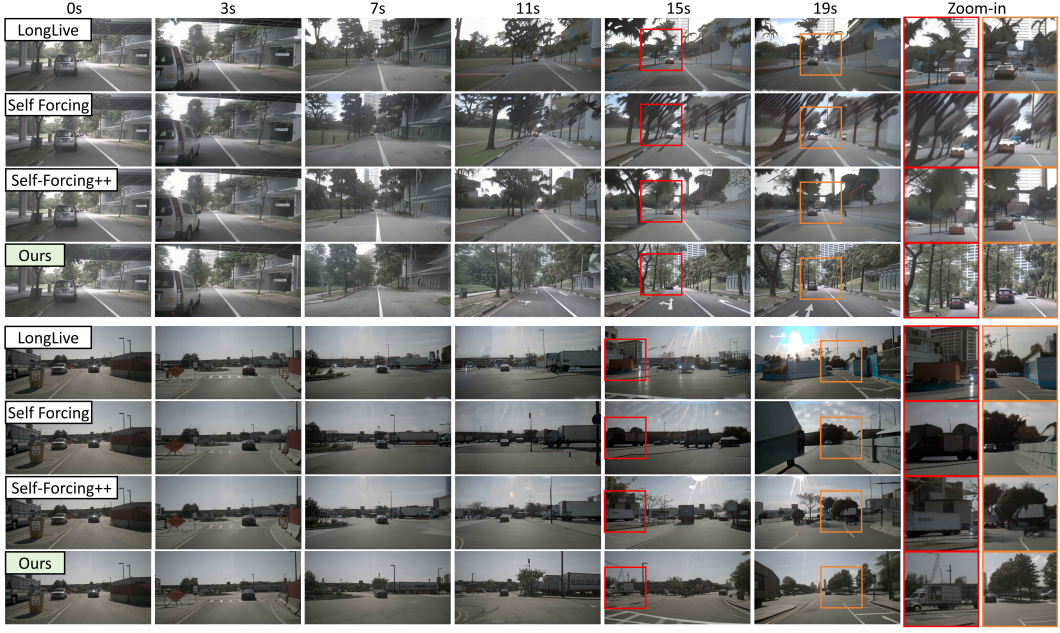}
    }
    \vspace{-2ex}
    \caption{\textbf{Long-horizon rollout comparison with streaming video generation methods.} Our method preserves clearer scene structure, more stable object geometry, and better visual quality over time. See ``zoom-in'' for better details.}
    \label{fig:res1}
\end{figure}

\begin{table}[t]
\centering
\caption{
    \textbf{Comparison with long-horizon autoregressive video generation baselines on nuScenes val.}
    \emph{Top group}: end-to-end interactive world model frameworks; they do not natively accept our driving control signals, so spatio-temporal consistency (ARE/DTW) is not applicable (``N/A'').
    \emph{Bottom group}: general streaming video generation recipes re-trained on our base model and data with our driving control modules, where all metrics are reported.
    Best results are in \textbf{bold}.
}
\label{tab:main_comparison}
\setlength{\tabcolsep}{3pt}
\renewcommand{\arraystretch}{0.95}
\begin{tabular*}{0.95\linewidth}{@{\extracolsep{\fill}}l cc cccc cc@{}}
\toprule
\multirow{2}{*}{Method} & \multicolumn{2}{c}{Visual} & \multicolumn{3}{c}{VBench} & \multicolumn{2}{c}{ST Cons.} \\
\cmidrule(lr){2-3} \cmidrule(lr){4-6} \cmidrule(lr){7-8}
 & FID$\downarrow$ & FVD$\downarrow$ & Qual.$\uparrow$ &  Mot.$\uparrow$ & Img.$\uparrow$ & ARE$\downarrow$ & DTW$\downarrow$ \\
\midrule
\multicolumn{8}{l}{\emph{Long-horizon interactive world model frameworks}} \\
Matrix-Game3~\citep{wang2026matrix}        & 35.69 & 338.22 & 78.99 & 93.78 & 60.44 & N/A & N/A \\
Helios~\citep{yuan2026helios}             & 30.53 & 218.23 & 79.02 &  95.03 & 58.82 & N/A & N/A \\
Causal-Forcing~\citep{zhu2026causal}      & 49.07 & 373.29 & 74.35 & 92.42 & 59.00 & N/A & N/A \\
HY-WorldPlay~\citep{sun2025worldplay}       & 33.51 & 580.72 & 76.58 & 99.48 & 58.60 & N/A & N/A \\
LingBot-World~\citep{team2026advancing}      & 37.67 & 325.55 & 77.08 & 92.87 & 55.55 & N/A & N/A \\
\midrule
\multicolumn{8}{l}{\emph{Long-horizon streaming video generation methods (re-trained on our base model and data)}} \\
Self-Forcing~\citep{huang2025self}    & 41.53 & 161.00 & 79.27 & 94.17 & 59.65 & 3.47 & 6.22 \\
Self-Forcing++~\citep{cui2025self}    & 28.84 & 147.57 & 79.47 & 93.92 & 60.25 & 3.78 & 3.61 \\
LongLive~\citep{yang2025longlive}     & 29.05 & 161.41 & 79.35 & 93.46 & 60.80 & 3.28 & 3.65 \\
\midrule
\method~(Ours)  & \textbf{13.82} & \textbf{92.99} & \textbf{79.53} & 93.85 & \textbf{62.50} & \textbf{2.60} & \textbf{3.27} \\
\bottomrule
\end{tabular*}
\end{table}

\begin{table}[t]
\centering
\caption{\textbf{Comparison with driving video generation methods on nuScenes val.}
Our distilled model achieves state-of-the-art FID and FVD both in the short-video and the long-video group.
}
\label{tab:driving_specific}
\setlength{\tabcolsep}{4pt}
\renewcommand{\arraystretch}{0.95}
\begin{tabular*}{0.95\linewidth}{@{\extracolsep{\fill}}l c c c c c@{}}
\toprule
Method & Unroll $N$ & Frames & Conditions & FID $\downarrow$ & FVD $\downarrow$ \\
\midrule
\multicolumn{6}{l}{\emph{Short-Video Evaluation}} \\
DriveDreamer~\citep{wang2024drivedreamer}     & 1 & 8   & T+M+B   & 14.90 & 340.80 \\
Panacea~\citep{wen2024panacea}           & 1 & 8  & T+M+B   &  16.90 & 139.00 \\
DreamForge~\citep{mei2024dreamforge}     & 1 & 16  & T+M+B   &  14.61 & 103.61 \\
Vista~\citep{gao2024vista}               & 1 & 25  & T+A     &   6.90 &  89.40 \\
\method~(Ours)                            & 1 & 21  & T+M+B+A & 12.54 & 84.53 \\
\midrule
\multicolumn{6}{l}{\emph{Long-Video Evaluation}} \\
MagicDrive-V2~\citep{gao2025magicdrive}  & 1 & 241 & T+M+B+A & 20.91 & 94.84 \\
\method~(Ours)                            & 20 & 211 & T+M+B+A & 13.82 & 92.99 \\
\bottomrule
\end{tabular*}
\\
{\footnotesize Conditions: T = text, M = HD map, B = bounding boxes, A = ego action.}
\end{table}

\subsection{Ablation studies}
We ablate three design choices in TRD: (1)~whether the teacher and student are initialized from SRR-trained weights, (2)~the number of autoregressive rollout steps $N$ during training, and (3)~the CFG augmentation strategy.
The CFG threshold $\tau_{\mathrm{th}}$ controls the maximum noise level at which CFG is applied in the DMD loss; we compare four schedules: \textbf{None} ($\tau_{\mathrm{th}}{=}0$), \textbf{Full} ($\tau_{\mathrm{th}}{=}1000$ throughout), \textbf{Early} ($\tau_{\mathrm{th}}$ decays $1000{\to}0$ from step~0), and \textbf{Delayed} ($\tau_{\mathrm{th}}$ decays $1000{\to}0$ after a 100-step warmup).
Results are shown in~\cref{tab:ablation_trd}.

\paragraph{Rollout-capable initialization benefits both teacher and student.}
When neither side is initialized from SRR (row~1, Base/Base), the model struggles with autoregressive error accumulation, producing the worst FVD (141.88) in this initialization group.
Switching only the student to SRR (row~2, SRR/Base) yields a marginal gain on FVD and slightly worse FID (FID 19.24$\to$20.34, FVD 141.88$\to$128.77): a rollout-aware student alone cannot compensate for a teacher that drifts under its own AR rollouts during TRD, and distilling from such degraded trajectories yields a degraded student.
Switching only the teacher to SRR (row~3, Base/SRR) yields the largest single-factor gain (FID 19.24$\to$14.44, FVD 141.88$\to$107.54), confirming that \emph{long-horizon supervision is dominated by the teacher's rollout reliability}---a teacher that can reliably roll out provides high-quality trajectory targets even when the student starts from Base.
Combining both (last row, SRR/SRR with our default $N{=}20$ and Delayed CFG) yields the best overall result (FID 13.82, FVD 92.99), showing that SRR is beneficial at both ends of the distillation pipeline while the teacher side carries the dominant effect.

\paragraph{Longer rollout during training improves long-horizon quality.}
With $N{=}1$, the student is only supervised on single-chunk generation and never experiences its own rollout errors, resulting in the worst ARE (3.39$^\circ$) and DTW (5.42) in this group.
As $N$ grows, both ARE and DTW decrease monotonically (ARE: 3.39$^\circ\to$3.03$^\circ\to$2.60$^\circ$; DTW: 5.42$\to$3.66$\to$3.27), and FVD drops sharply once $N$ is large enough to expose the student to multi-step rollout errors (139.28 at $N{=}4$ $\to$ 92.99 at $N{=}20$).
This confirms that longer autoregressive training chains are essential for deployment-time stability.

\paragraph{CFG augmentation benefits from noise-truncation.}
Directly applying CFG at all noise levels (Full) causes severe oversaturation, nearly doubling FVD to 184.06 compared to 110.81 without CFG.
Early decay yields little improvement over no CFG, because DMD training itself requires few steps and the high-noise CFG is phased out before it can take effect.
Delayed decay resolves this by allowing a warmup period for full-range CFG to establish strong conditional generation, then progressively restricting CFG to low noise levels to maintain visual quality, achieving the best FID (13.82), FVD (92.99), and ARE (2.60$^\circ$); the Early schedule reaches a marginally better DTW (3.13 vs.\ 3.27) but at a sizeable cost in visual fidelity (FID 14.70 and FVD 111.99), and we therefore adopt Delayed decay as the default.

\begin{table}[t]
\centering
\caption{
    \textbf{Ablation of TRD design choices on nuScenes val.}
    Each group varies one factor while fixing the others.
    Init: student / teacher initialization (Base or SRR).
}
\label{tab:ablation_trd}
\setlength{\tabcolsep}{4pt}
\renewcommand{\arraystretch}{0.95}
\begin{tabular*}{0.95\linewidth}{@{\extracolsep{\fill}}l c l c c c c@{}}
\toprule
Init (Stu. / Tea.) & Unroll $N$ & CFG & FID$\downarrow$ & FVD$\downarrow$ & ARE$\downarrow$ & DTW$\downarrow$ \\
\midrule
Base / Base  & \multirow{3}{*}{20} & \multirow{3}{*}{Delayed} & 19.24 & 141.88 & 2.76 & 3.30 \\
SRR / Base  &    &         & 20.34 & 128.77 & 3.15 & 3.39 \\
Base / SRR  &    &         & \underline{14.44} & \underline{107.54} & 2.75 & 3.80 \\
\midrule
\multirow{2}{*}{SRR / SRR} & 1 & \multirow{2}{*}{Delayed} & 21.15 & 135.35 & 3.39 & 5.42 \\
      & 4 &                           & 18.19 & 139.28 & 3.03 & 3.66 \\
\midrule
\multirow{4}{*}{SRR / SRR} & \multirow{4}{*}{20} & None    & 14.59 & 110.81 & 2.77 & 3.28 \\
      &                       & Full    & 20.84 & 184.06 & 3.86 & 3.96 \\
      &                       & Early   & 14.70 & 111.99 & \underline{2.64} & \textbf{3.13} \\
      &                       & Delayed & \textbf{13.82} & \textbf{92.99} & \textbf{2.60} & \underline{3.27} \\
\bottomrule
\end{tabular*}
\end{table}

\section{Conclusion and Limitations}
\label{sec: limitations}

We presented \method, an anti-drifting distillation framework for long-horizon autonomous driving simulation. Instead of relying on frame sinks or student-side degradation alone, \method first turns a standard video diffusion model into a rollout-capable teacher through scheduled rollout recovery, and then distills its long-horizon AR behavior into a short-chunk student with teacher rollout DMD. This design provides reliable long-horizon supervision under bounded memory and enables efficient AR deployment.

A key limitation is that SRR is conducted offline; future work will explore online rollout-recovery training, where the world model continuously improves its AR robustness from its own interaction trajectories.

{\small
\bibliographystyle{plainnat}
\bibliography{reference}
}

\clearpage
\newpage
\appendix

\section{Implementation details}
\label{appx:implementation}

\paragraph{Backbone and VAE.}
\method\ is built on Wan~2.1~1.3B~\citep{wan2025wan} with full bidirectional attention, and adopts the disentangled driving-control modules described in~\cref{subsec: Base model training}. Since driving scenes involve fast ego-motion and rapidly changing fine details, we fine-tune the original VAE to reduce its temporal compression ratio from 4 to 1, preserving full temporal resolution for higher visual fidelity.

\paragraph{Conditional control modules.}
\label{appx:control_modules}
We elaborate on the disentangled control architecture summarized in~\cref{subsec: Base model training}. HD map and bounding box information are first rendered as structural conditions $z_{\bm{b}_{f}} \in \mathbb{R}^{c \times f \times h \times w}$, processed by a lightweight convolutional adapter, and reshaped into layout tokens $\bm{h}_{\bm{b}_{f}} \in \mathbb{R}^{f \times s \times d}$. The evolving feature representation at diffusion time $t$, denoted $\bm{h}_{(t)}$, is updated by adding the projected layout tokens following $\bm{h}_{(t)} \leftarrow \bm{h}_{(t)} + f_{\text{zero}}\!\left(\bm{h}_{\bm{b}_{f}}\right)$, where $f_{\text{zero}}(\cdot)$ is a zero-initialized projector.
Action control is injected by computing the relative transformation $\Delta \mathbf{T}_i = \mathbf{T}_{i}^{-1}\mathbf{T}_{i+1}$ from a continuous trajectory of shape $F \times 4 \times 4$. We extract the residual signal $\bm{a} = (\Delta x, \Delta y, \Delta \text{yaw}) \in \mathbb{R}^{F \times 3}$ and inject it via AdaLN-style gating~\citep{peebles2023scalable} as $f_{\text{zero}}\!\left(\phi(\bm{a})\right) \in \mathbb{R}^{f \times 6 \times d}$, where $\phi(\cdot)$ is the sinusoidal frequency embedding. The resulting six channels are split into two groups, representing the pre-normalization shift/scale and post-layer residual gate for self-attention and feed-forward sublayers, respectively.


\paragraph{Training stages.}
We first train the base conditional driving world model $\mathcal{G}_0$ for 40K steps on a 100-hour proprietary driving dataset and 10K steps on nuScenes~\citep{caesar2020nuscenes}, using short clips of length $T{+}K$ with ground-truth conditioning latents.
We then apply SRR for 10K steps on nuScenes to obtain $\mathcal{G}_{\mathrm{roll}}$; SRR keeps the same $(T, K)$ but maintains a per-clip rollout cache refreshed every $R$ optimizer steps with the current $\theta$, uses a decaying AR rollout depth $N(k)$, samples the boundary index $s$ in $[T,\, T{+}N(k) K]$ within each clip, and gradually enlarges the blending radius $w$.
For TRD distillation, the SRR-initialized teacher and student share a context window of $T{=}11$ frames; the teacher generates $K^{\mathcal{T}}{=}40$ frames per chunk and slides a $(T{+}K^{\mathcal{T}}){=}51$-frame supervision window across its rollout to cover each student chunk, providing memory-bounded supervision over the long-horizon trajectory, while the student generates $K^{\mathcal{S}}{=}10$ new frames per chunk with $M^{\mathcal{S}}{=}4$ denoising steps and rolls out $N{=}20$ autoregressive steps. TRD updates the chunk-wise distribution matching loss every $D$ student chunks; CFG augmentation is applied only at low noise levels ($\tau \le \tau_{\mathrm{th}}$) with weight $\alpha$.
Full optimization hyperparameters and stage-specific schedules are listed in~\cref{tab:training_config_teacher,tab:training_config_distill}.

\paragraph{Datasets and evaluation horizon.}
For nuScenes~\citep{caesar2020nuscenes}, we use 700 multi-view videos of $\sim$20 seconds for training and 150 for validation; the per-clip length of $\sim$20\,s is the dataset upper bound, which determines the horizon of our quantitative evaluation. Longer rollouts of up to one minute on a self-collected dataset are demonstrated qualitatively in~\cref{appx:minute}.

\paragraph{Inference efficiency.}
With our temporally-uncompressed VAE and a 4-step student denoiser, each AR rollout step generates one $K^{\mathcal{S}}{=}10$-frame chunk on a single NVIDIA 5090 GPU. The per-chunk wall-clock latency is \textbf{1.8\,s} at $256{\times}512$ resolution and \textbf{5.8\,s} at $384{\times}768$, yielding effective generation rates of $\sim$5.6\,FPS and $\sim$1.7\,FPS, respectively. Because the rolling-window rollout (\cref{subsec: Base model training}) bounds per-step compute by $T{+}K^{\mathcal{S}}$ regardless of total horizon, this per-chunk latency is sustained throughout the rollout, enabling stable streaming generation up to minute-scale horizons (\cref{appx:minute}).

\section{Evaluation metrics}
\label{appx:metrics}

\paragraph{Visual quality.}
We report Fr\'{e}chet Inception Distance (FID) and Fr\'{e}chet Video Distance (FVD) computed over the full rollout sequence, as well as VBench~\citep{huang2024vbench} as an overall video-quality measure.

\paragraph{Spatio-temporal consistency.}
These metrics are applicable only to methods that accept driving control signals. We recover per-frame camera poses from both generated and ground-truth videos with VGGT~\citep{wang2025vggt}. \textbf{ARE} (Average Rotation Error) measures the mean geodesic distance between predicted and GT rotation matrices across frames, reflecting heading accuracy. \textbf{DTW} (Dynamic Time Warping)~\citep{keogh2000scaling} aligns predicted and GT ego-motion trajectories via non-rigid time warping and computes the cumulative Euclidean distance under the optimal alignment, capturing path-shape fidelity even under temporal misalignment.

\begin{table}[t]
\label{tab:training_config1}
    \centering
    \small
    \caption{\textbf{Training configuration of the teacher stages of \method.}
    ``--'' marks entries not applicable to the corresponding stage.}
    \label{tab:training_config_teacher}
    \setlength{\tabcolsep}{6pt}
    \begin{tabular}{l c c}
    \toprule
    Hyperparameter & Base ($\mathcal{G}_0$) & SRR ($\mathcal{G}_{\mathrm{roll}}$) \\
    \midrule
    Optimizer        & AdamW & AdamW \\
    Learning rate    & 1e-5  & 1e-5  \\
    Weight decay     & 1e-2   & 1e-5  \\
    Global batch size  & 96 & 64  \\
    Mixed precision  & bf16  & bf16  \\
    Training steps   & 40K (proprietary) + 10K (nuScenes) & 10K (nuScenes) \\
    GPU Usage   & 96 NVIDIA 5090  & 64 NVIDIA 5090 \\
    Context window length $T$  & 11 & 11 \\
    Chunk size $K$             & 10, 40 & 10, 40 \\
    Resolution & [256, 512], [384, 768] & [256, 512], [384, 768]\\
    AR rollout depth $N$       & -- & $N(k)$ schedule \\
    Cache refresh period $R$ (optimizer steps) & -- & 2000 \\
    $N(k)$ schedule (start $\to$ end)           & -- & 10 $\to$ 4 (steps 0-8000) \\
    Blending radius $w$ schedule               & -- & $0 \to 8$ (steps 0-8000) \\
    \bottomrule
    \end{tabular}
\end{table}


\begin{table}[t]
\label{tab:training_config2}
    \centering
    \small
    \caption{\textbf{Training configuration of the distillation stage (TRD).}}
    \label{tab:training_config_distill}
    \setlength{\tabcolsep}{6pt}
    \begin{tabular}{l c}
    \toprule
    Hyperparameter & TRD \\
    \midrule
    Optimizer        & AdamW \\
    Student learning rate    & 2e-6  \\
    Critic learning rate    & 1e-5  \\
    Weight decay     & 0.1  \\
    Batch size       & 32  \\
    Mixed precision  & bf16  \\
    Training steps   & 200  \\
    GPU Usage & 32 NVIDIA 5090 \\
    Context window length $T$  & 11 \\
    Chunk size $K$             & 10 (student) / 40 (teacher) \\
    AR rollout depth $N$       & 20 (student) \\
    Denoising steps            & 4 \\
    DMD update interval $D$ (student chunks)   & 5 \\
    CFG threshold $\tau_{\mathrm{th}}$ schedule & 1000 (steps 0–100), decay $\rightarrow$ 0 (step 400) \\
    CFG scale $\alpha$                         & 6 \\
    \bottomrule
    \end{tabular}
\end{table}

\section{Baseline evaluation protocols}
\label{appx:baselines}

\paragraph{Group (i): long-horizon interactive world model frameworks.}
These methods are designed for general open-domain or interactive world simulation and do not natively accept our driving control signals (actions, HD maps, bounding boxes). For each method, we use the publicly released checkpoint, condition it only on the initial frame(s) following the original interface, and let it autoregressively roll out to match our nuScenes evaluation horizon. We report visual-quality metrics (FID, FVD, VBench) only; ego-motion consistency (ARE/DTW) is marked ``N/A'' since no driving control is provided.

\paragraph{Group (ii): long-horizon streaming video generation methods.}
Self-Forcing~\citep{huang2025self}, Self-Forcing++~\citep{cui2025self}, and LongLive~\citep{yang2025longlive} are general streaming long-video recipes. To ensure fair comparison, for each baseline we (a) attach our disentangled driving control modules (\cref{subsec: Base model training}) to the same Wan~2.1~1.3B backbone, (b) initialize from our base model $\mathcal{G}_0$, and (c) re-train using each method's original training recipe (loss, scheduler, rollout configuration, hyperparameters) on our nuScenes training split. Architecture, base model, training data, and driving conditioning are therefore shared with \method, so the only remaining variable is the long-horizon training framework itself. All metrics (FID, FVD, VBench, ARE, DTW) are reported under the same protocol as \method.

\paragraph{Group (iii): driving video generation models.}
These are domain-specific driving generators that accept driving control signals but produce only a fixed-length single-pass clip (8--241 frames depending on the method). Since each method's evaluation setting differs (clip length, validation split, conditioning, evaluation protocol), we directly take FID/FVD numbers from the original papers and report them in two groups---\emph{short-video} (8--25 frames) and \emph{long-video} (241 frames)---rather than re-running them under a unified protocol. Results in this table are intended as a domain reference rather than a strict head-to-head ranking.

\begin{figure}
    \centering
    \includegraphics[width=0.5\linewidth]{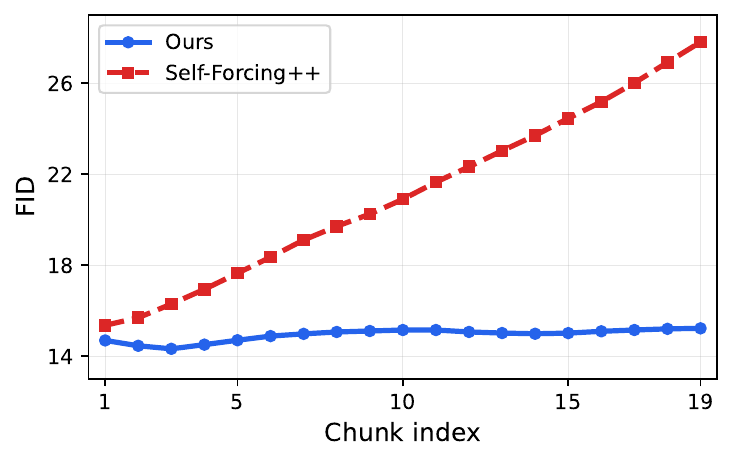}
    \caption{Long-horizon generation quality comparison.}
    \label{fig:fid_per_chunk}
\end{figure}

\section{Error accumulation analysis} 
Fig.~\ref{fig:fid_per_chunk} reports the FID computed on each cumulative chunk during autoregressive rollout. Our method maintains stable image quality across all 19 chunks, with no accumulation of error over the rollout horizon. In contrast, Self-Forcing++ suffers from compounding errors that cause FID to degrade monotonically, confirming that our self-rollout recovery training yields significantly stronger robustness to the distribution shift encountered during long-horizon autoregressive generation.

\section{Additional qualitative results on nuScenes}
\label{appx:nusc_qual}

\begin{figure}[t]
    \centering
    \makebox[\linewidth][c]{
        \includegraphics[width=0.95\linewidth]{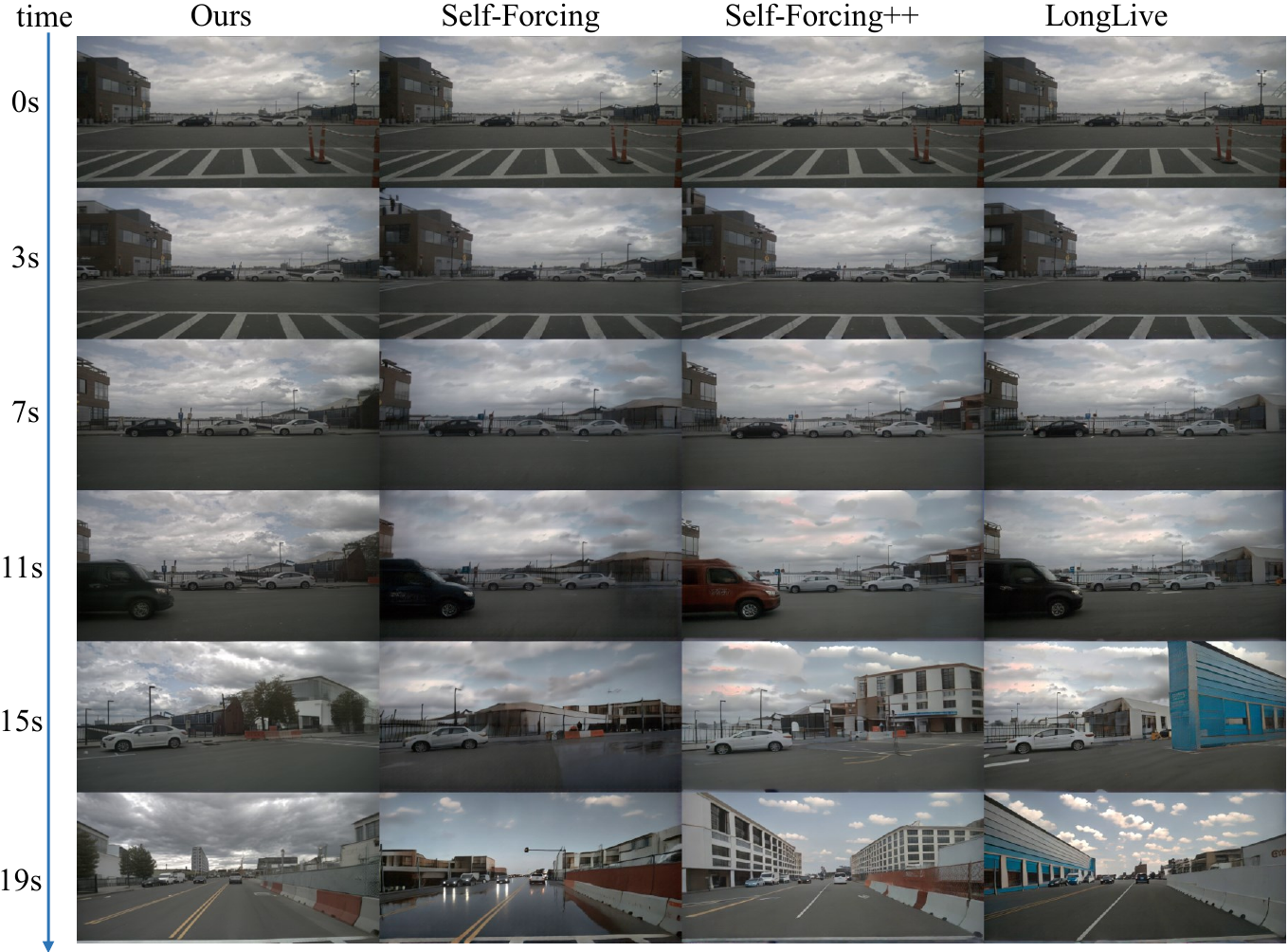}
    }
    \caption{\textbf{Qualitative comparison with long-horizon streaming baselines on nuScenes val (scene 1).} From left to right: \method, Self-Forcing, Self-Forcing++, LongLive}
    \label{fig:sup_nusc_baseline_1}
\end{figure}

\begin{figure}[t]
    \centering
    \makebox[\linewidth][c]{
        \includegraphics[width=0.95\linewidth]{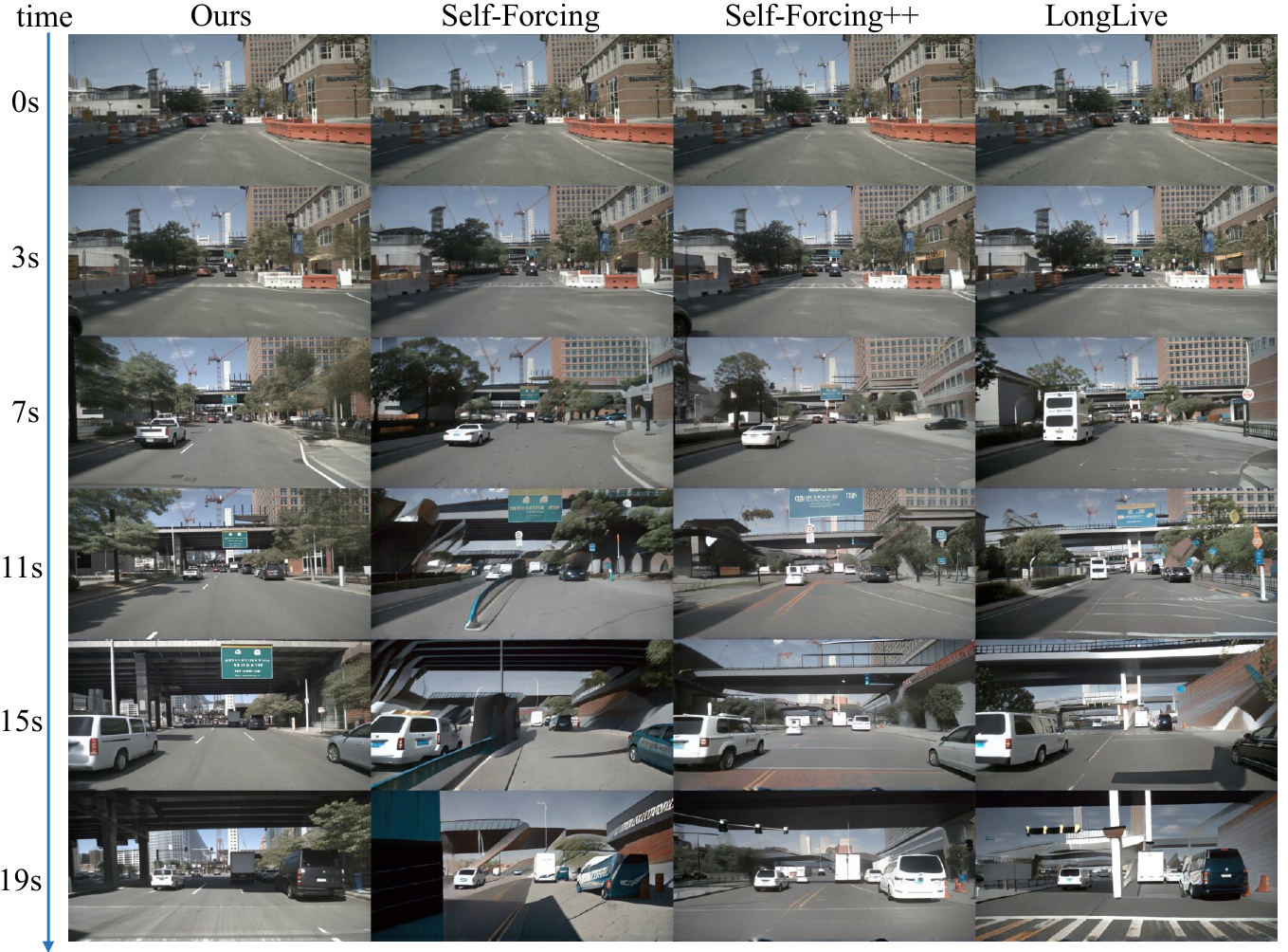}
    }
    \caption{\textbf{Qualitative comparison with long-horizon streaming baselines on nuScenes val (scene 2).}}
    \label{fig:sup_nusc_baseline_3}
\end{figure}

\begin{figure}[t]
    \centering
    \makebox[\linewidth][c]{
        \includegraphics[width=0.95\linewidth]{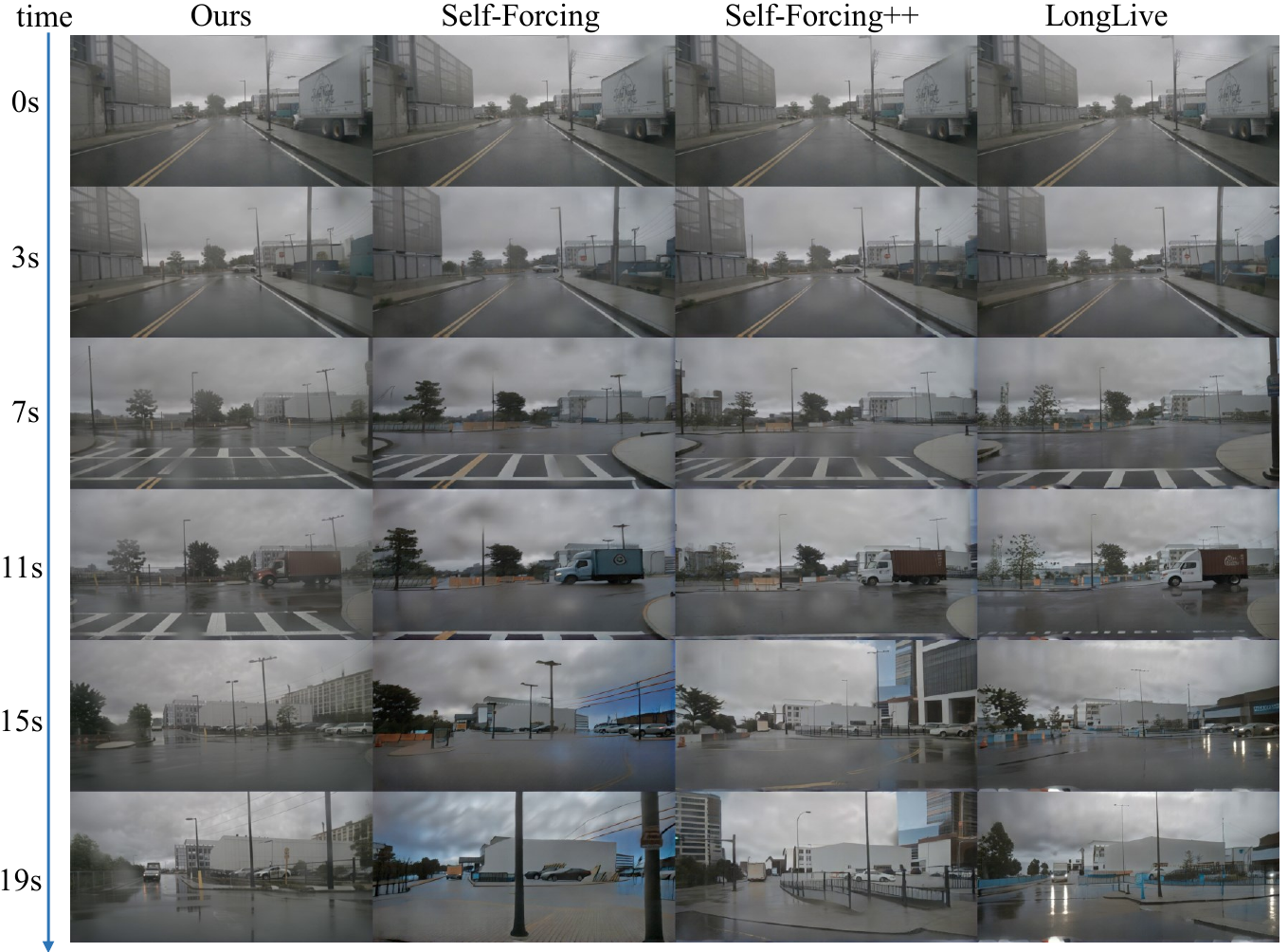}
    }
    \caption{\textbf{Qualitative comparison with long-horizon streaming baselines on nuScenes val (scene 3).}
    Same layout as~\cref{fig:sup_nusc_baseline_1}.}
    \label{fig:sup_nusc_baseline_4}
\end{figure}

\begin{figure}[t]
    \centering
    \makebox[\linewidth][c]{
        \includegraphics[width=0.95\linewidth]{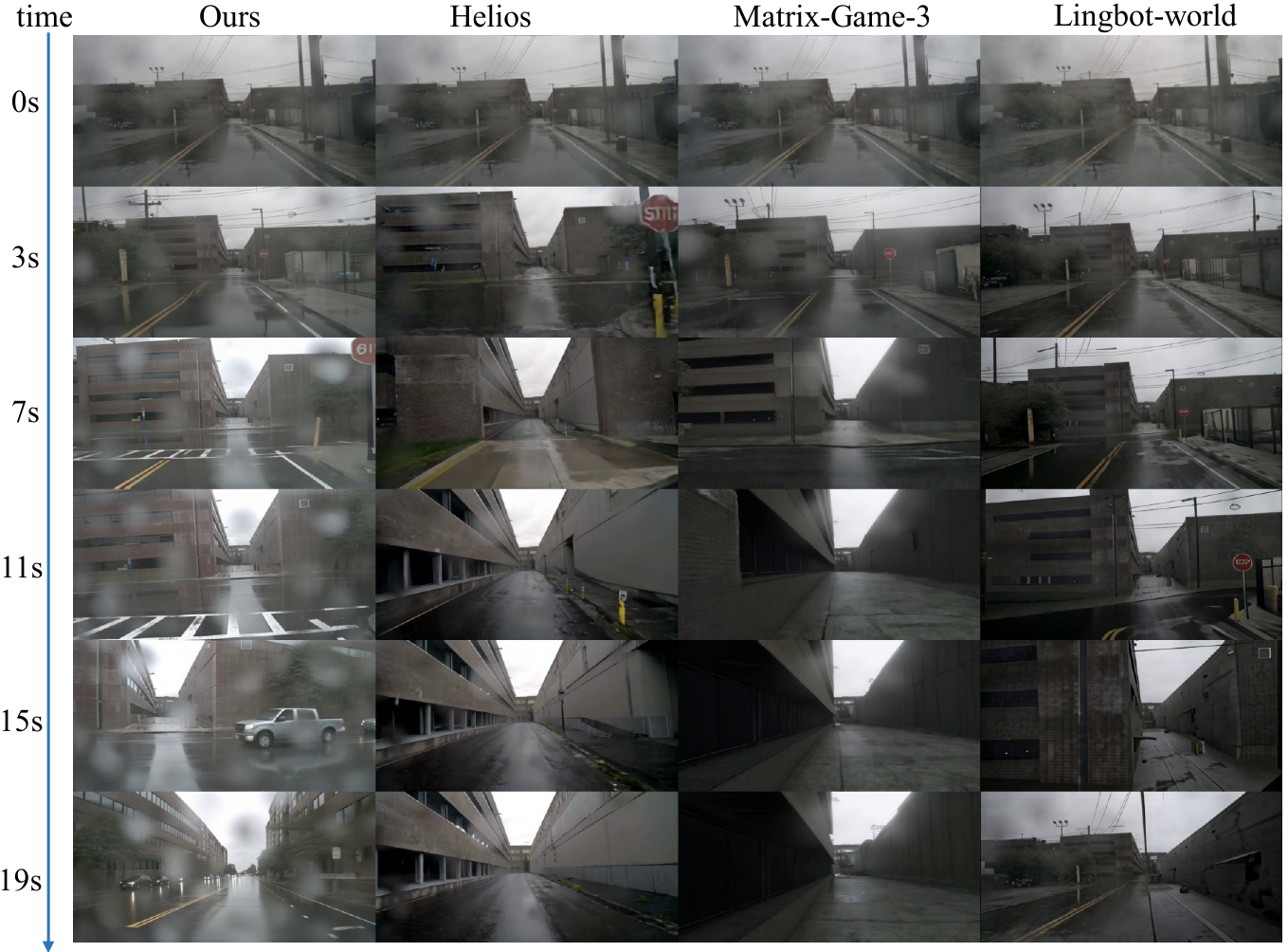}
    }
    \caption{\textbf{Qualitative comparison with driving world models on nuScenes val (scene 1).}
    From left to right: \method, Helios, Matrix-Game3, LingBot-World}
    \label{fig:sup_nusc_wm_1}
\end{figure}

\begin{figure}[t]
    \centering
    \makebox[\linewidth][c]{
        \includegraphics[width=0.95\linewidth]{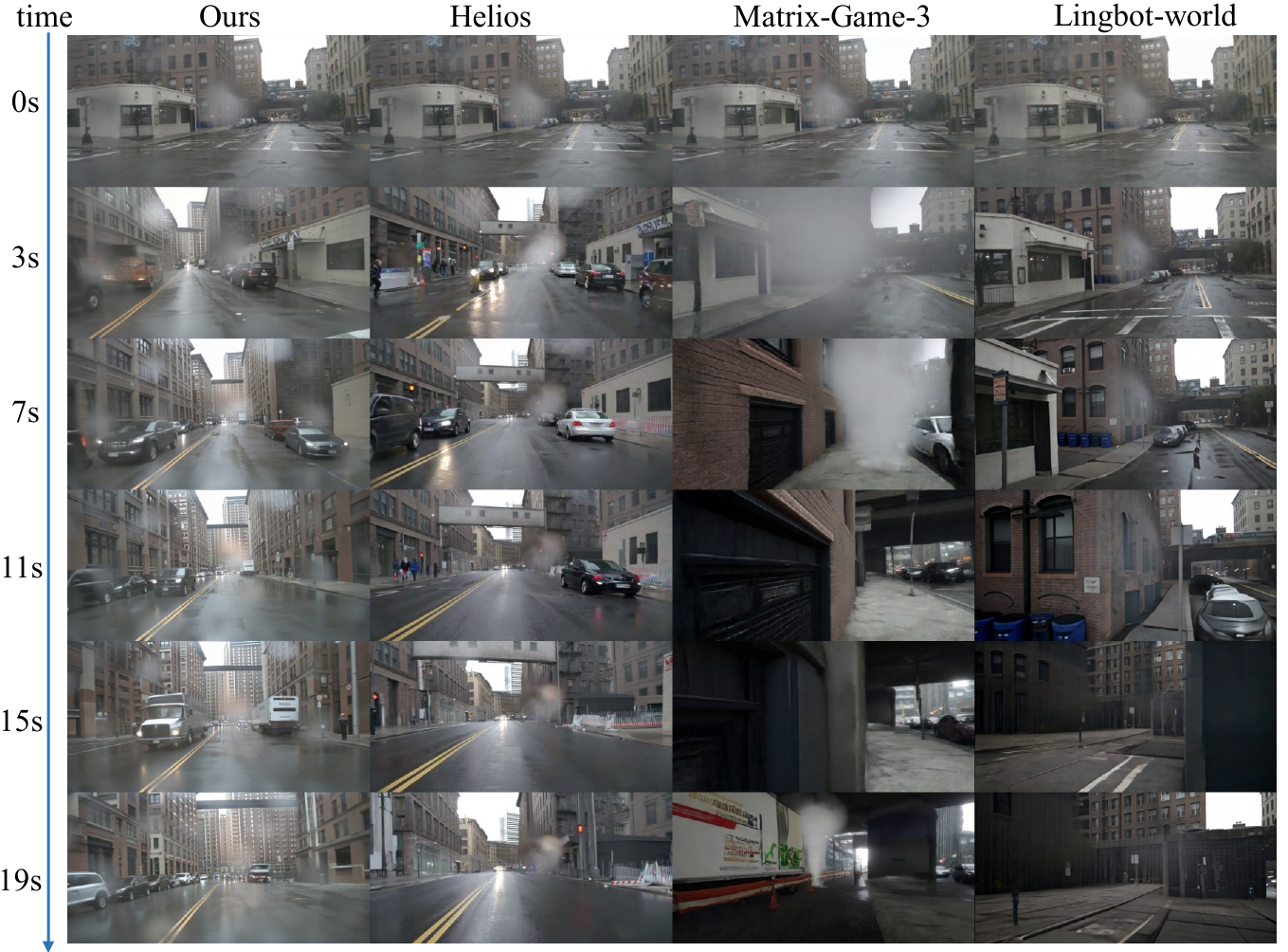}
    }
    \caption{\textbf{Qualitative comparison with driving world models on nuScenes val (scene 2).}}
    \label{fig:sup_nusc_wm_2}
\end{figure}

We provide additional rollouts on nuScenes val to complement the quantitative comparison in the main paper.
\Cref{fig:sup_nusc_baseline_1,fig:sup_nusc_baseline_3,fig:sup_nusc_baseline_4} compare \method against the long-horizon streaming baselines (Self-Forcing, Self-Forcing++, LongLive) under our shared base model and driving control conditions.
\Cref{fig:sup_nusc_wm_1,fig:sup_nusc_wm_2} compare against representative driving world models that are not autoregressive (e.g., Matrix-Game3, Helios, LingBot-World), where each method uses its own native generation setting.
Across both groups, \method preserves road structure, agent layout, and spatio-temporal consistency throughout the rollout, while baselines progressively drift in geometry, color, or scene composition.

\section{Results on the self-collected dataset}
\label{appx:e2e}

\begin{figure}[t]
    \centering
    \makebox[\linewidth][c]{
        \includegraphics[width=0.95\linewidth]{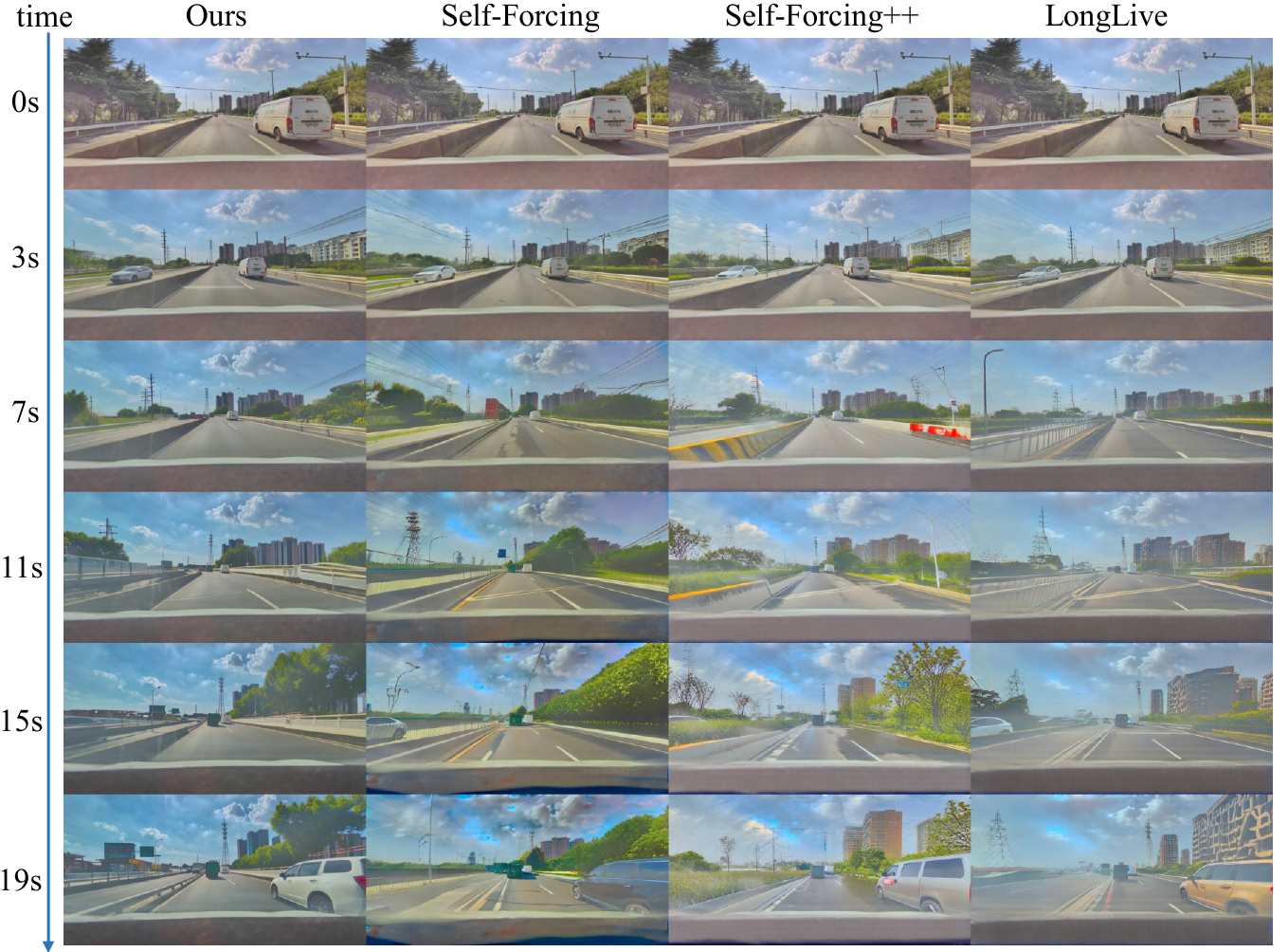}
    }
    \caption{\textbf{Qualitative comparison with long-horizon streaming baselines on the self-collected (e2e) dataset (scene 1).} From left to right: \method, Self-Forcing, Self-Forcing++, LongLive}
    \label{fig:sup_e2e_1}
\end{figure}

\begin{figure}[t]
    \centering
    \makebox[\linewidth][c]{
        \includegraphics[width=0.95\linewidth]{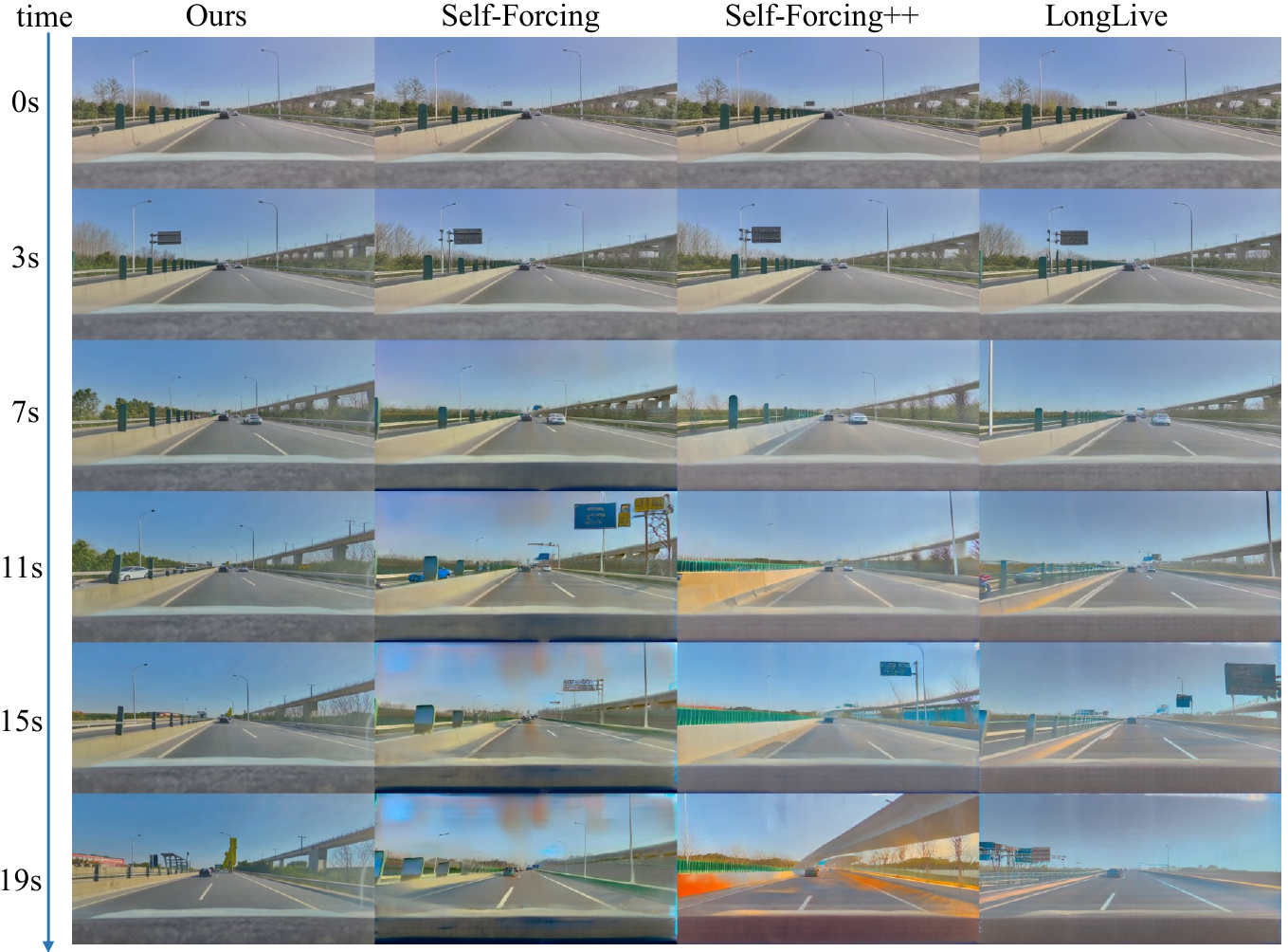}
    }
    \caption{\textbf{Qualitative rollouts on the self-collected (e2e) dataset (scene 2).}}
    \label{fig:sup_e2e_2}
\end{figure}

To evaluate generalization beyond nuScenes, we also benchmark on our internal self-collected end-to-end (e2e) driving dataset, which features higher ego speeds and more diverse scenarios than nuScenes.
\Cref{tab:e2e_comparison} reports the quantitative comparison against the long-horizon streaming baselines under the same protocol as the main paper (re-trained on our base model and the e2e training split, evaluated on a held-out split).
\Cref{fig:sup_e2e_1,fig:sup_e2e_2} provide qualitative rollouts.
\method again outperforms all baselines on visual quality (FID/FVD), VBench dimensions, and spatio-temporal consistency, indicating that the rollout-capable base model and TRD distillation transfer to a substantially different driving distribution.

\begin{table}[t]
\centering
\caption{
    \textbf{Comparison with long-horizon streaming video generation baselines on the self-collected (e2e) dataset.}
    All baselines share our base model and driving control modules, and are re-trained on the e2e training split following each method's original recipe.
    Best results are in \textbf{bold}; second best are \underline{underlined}.
}
\label{tab:e2e_comparison}
\setlength{\tabcolsep}{3pt}
\begin{tabular}{l cc cccc cc}
\toprule
\multirow{2}{*}{Method} & \multicolumn{2}{c}{Visual} & \multicolumn{3}{c}{VBench} & \multicolumn{2}{c}{ST Cons.} \\
\cmidrule(lr){2-3} \cmidrule(lr){4-6} \cmidrule(lr){7-8}
 & FID$\downarrow$ & FVD$\downarrow$ & Qual.$\uparrow$ &  Mot.$\uparrow$ & Img.$\uparrow$ & ARE$\downarrow$ & DTW$\downarrow$ \\
\midrule
\multicolumn{8}{l}{\emph{Long-horizon streaming video generation methods (re-trained on our base model and e2e data)}} \\
Self-Forcing~\citep{huang2025self}    & 58.23 & 561.11 & 76.68 & 94.48  & \underline{63.18}  & 5.43 & 14.13 \\
Self-Forcing++~\citep{cui2025self}    & 66.93 & 534.36 & 74.54 & 92.70 & 59.12 & 7.32 & 18.40 \\
LongLive~\citep{yang2025longlive}     & \underline{28.39} & \underline{374.94} & \underline{78.18} & \underline{94.57} & 62.53 & \underline{4.05} & \underline{8.11} \\
\midrule
\method~(Ours)  & \textbf{12.01} & \textbf{117.27} & \textbf{80.12} & \textbf{95.22} & \textbf{67.65} & \textbf{3.67} & \textbf{5.29} \\
\bottomrule
\end{tabular}
\end{table}

\section{Minute-level autoregressive generation}
\label{appx:minute}

\begin{figure}[t]
    \centering
    \makebox[\linewidth][c]{
        \includegraphics[width=\linewidth]{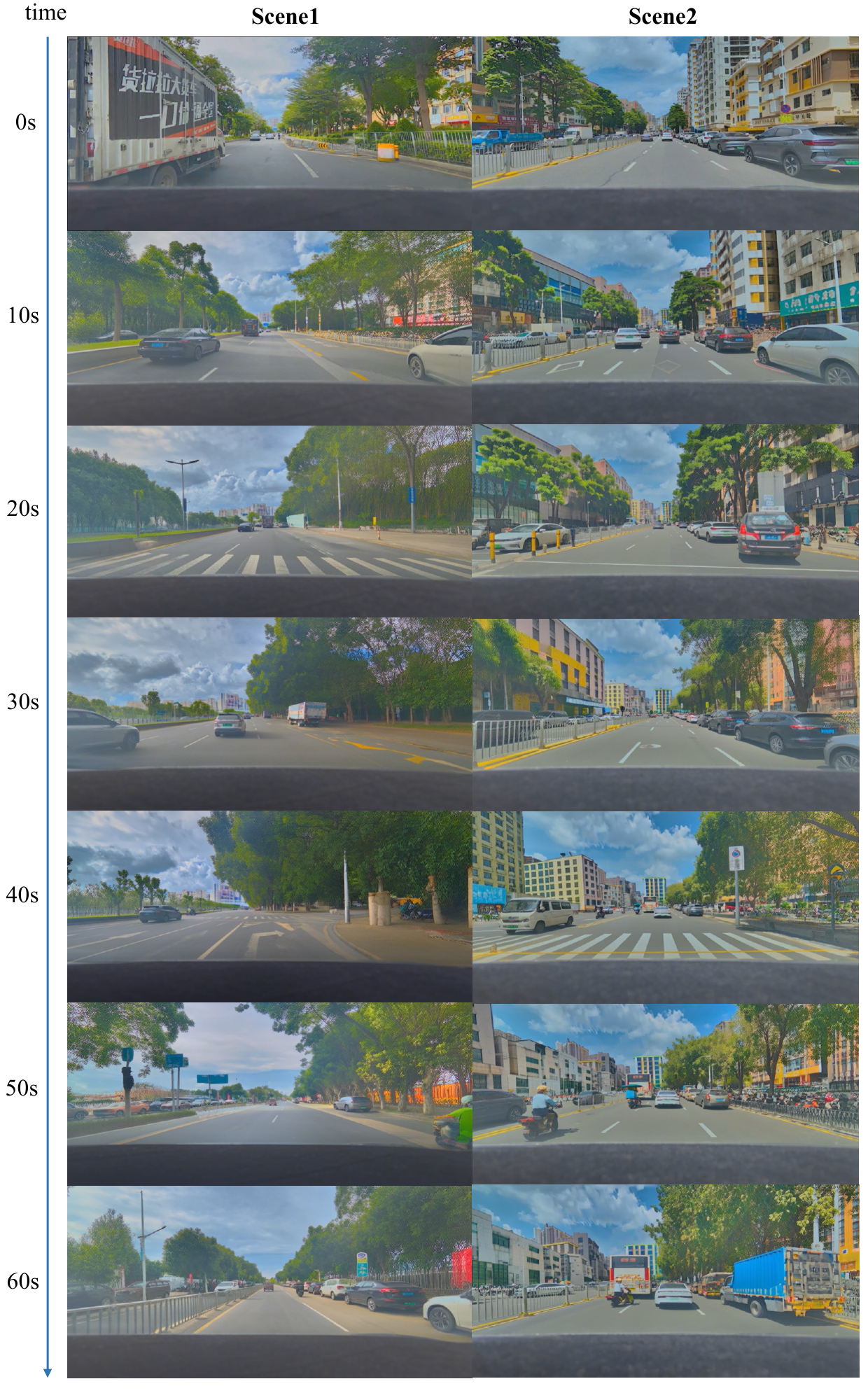}
    }
    \caption{\textbf{Minute-level autoregressive generation on the self-collected (e2e) dataset.}}
    \label{fig:sup_e2e_long}
\end{figure}

\method supports minute-level AR generation under bounded per-step memory and compute, which is otherwise infeasible for single-pass video diffusion baselines whose memory grows quadratically with sequence length.
At inference, the sliding-window rollout (\cref{subsec: Base model training}) keeps each step's compute bounded by $T{+}K^{\mathcal{S}}$, so the model can be rolled out indefinitely without re-warming the cache.
\Cref{fig:sup_e2e_long} shows a continuous rollout of approximately one minute on the self-collected dataset (10\,FPS, $K^{\mathcal{S}}{=}10$, $T{=}11$), where the model maintains coherent road geometry, lane structure, and traffic-agent behavior throughout the rollout, demonstrating practical viability for closed-loop driving simulation.

\section{Closed-loop driving simulation}
\label{appx:closed_loop}

\begin{figure}[t]
    \centering
    \makebox[\linewidth][c]{
        \includegraphics[width=\linewidth]{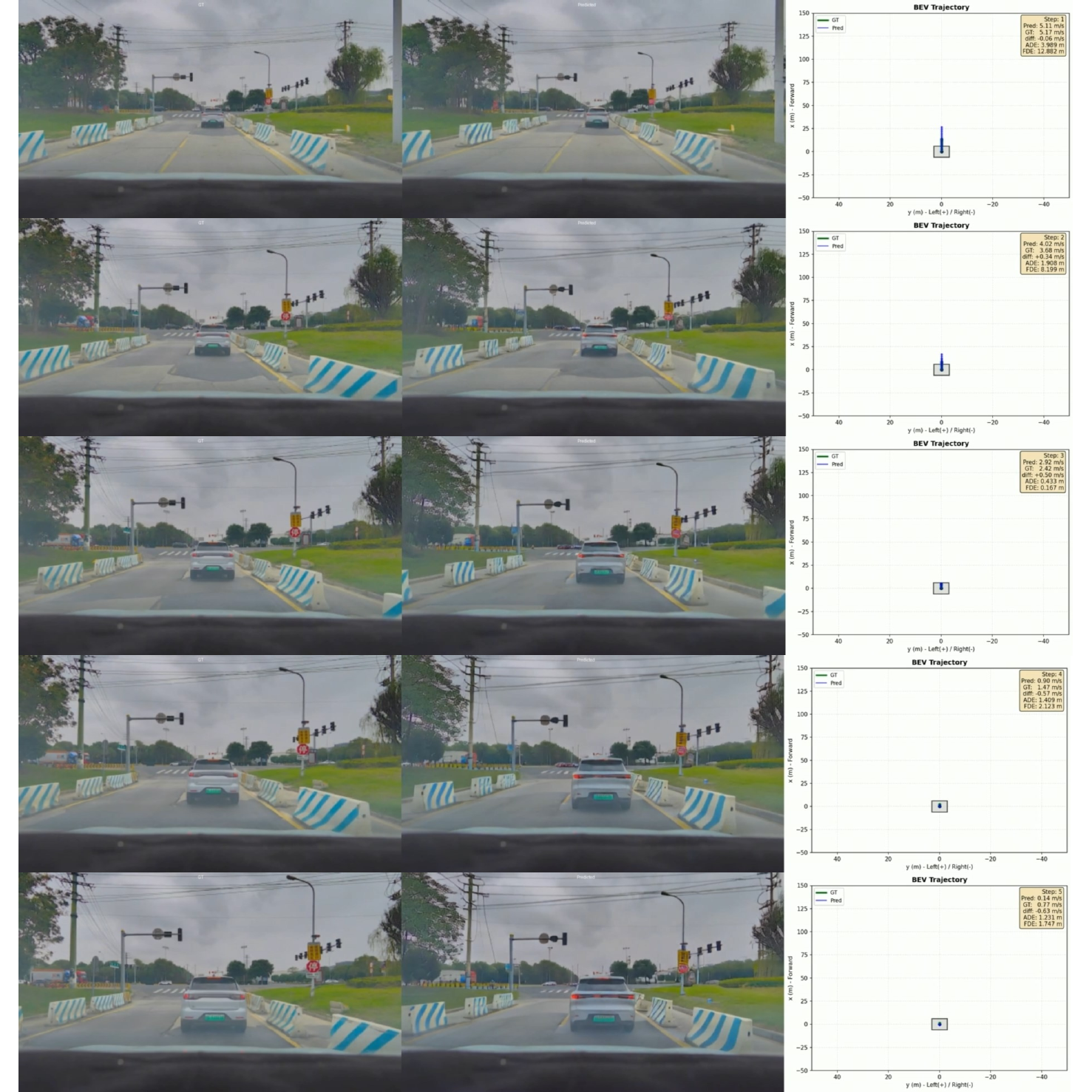}
    }
    \caption{\textbf{Closed-loop driving simulation.} A planner consumes the latest generated frame at each step and outputs an ego trajectory, which \method\ uses as the next-step action condition. Despite the planner-and-world-model loop being driven entirely by self-generated signals, \method\ maintains coherent scene structure and stable agent behavior over long horizons.}
    \label{fig:closed_loop}
\end{figure}

We further evaluate \method\ in a closed-loop driving simulation setup, where a planner and the world model interact step by step: at each AR step, the planner consumes the latest generated frame and outputs an ego trajectory, which is re-encoded (together with the corresponding HD map and bounding-box layouts) as the next-step action condition fed back into \method, and the loop repeats indefinitely under the same bounded per-step memory budget as standard rollout (\cref{appx:minute}). No ground-truth ego trajectory is used during simulation, so every conditioning signal beyond the initial frame is self-generated. As shown in~\cref{fig:closed_loop}, despite the compounding of rollout error from the world model and prediction error from the planner, \method\ maintains coherent road geometry, lane topology, and stable agent behavior throughout the rollout, demonstrating practical viability for closed-loop policy evaluation under self-generated ego trajectories.

\clearpage
\newpage

\end{document}